# Quality of Geographic Information: Ontological approach and Artificial Intelligence Tools in the *REV!GIS* project

Author: Robert Jeansoulin[(1)]

(1) LSIS/CMI, Université de Provence, 39 rue Joliot Curie, 13453 Marseille Cedex 13, France
Tel: 33-491 113 608 - Fax: 33-491 113 602 - <robert.jeansoulin@cmi.univ-mrs.fr>

**Abstract**:
The objective is to present one important aspect of the European IST-FET project "*REV!GIS*"[1]: the methodology which has been developped for the translation (interpretation) of the quality of the data into a "fitness for use" information, that we can confront to the user needs in its application.
This methodology is based upon the notion of "ontologies" as a conceptual framework able to capture the explicit and implicit knowledge involved in the application. We do not address the general problem of formalizing such ontologies, instead, we rather try to illustrate this with three applications which are particular cases of the more general "data fusion" problem. In each application, we show how to deploy our methodology, by comparing several possible solutions, and we try to enlighten where are the quality issues, and what kind of solution to privilegy, even at the expense of a highly complex computational approach. The expectation of the REV!GIS project is that computationally tractable solutions will be available among the next generation AI tools.

## 1. Introduction

The objective of the *REV!GIS* European project is to improve the use of the geographic information, in particular, uncertain, imprecise and incomplete information, by taking advantage of any kind of knowledge about data specifications and limits, and demonstrating that it is important to raise awareness on the information quality.

A commonly shared feeling is that the "metadata" delivered by the data producers are not useful for answering the question of the "fitness for use" of the data. Two major results of the project are: (1) the description and distinction of the data specifications and of the user requirements into separate ontologies greatly helps to understand the respective roles of the metadata and the user-required qualities; (2) Artificial Intelligence provides potential and sometimes effective tools, to manage the quality, and perhaps to "revise" parts of the geographic information detected as probably incorrect.

In any GIS application the user (the "modeller") constructs a reasoning model to interpret parts of the Real World, based on the concepts she wants to represent, and how she understands them, in order to solve her problem: let's name this the " Problem Ontology". But she also uses one or more data sets, which have different ontologies, that we may call "Product Ontologies". The correspondences between these ontologies may well be less than straight-forward. For example, different land-cover maps might use the same term 'bog' to represent one of their classes, but the precise meaning of this class will differ according to the product ontology: if it was based on satellite data, or a field survey, what algorithms were used, what training data etc. To be able to say something sensible about data quality one must understand something of the relationships between these ontologies.

---

[1] Website: http://www.cmi.univ-mrs.fr/REVIGIS/Full/

Ideally we'd like to embed metadata on data quality and quality of correspondences between concepts in a formal language: essentially generating a formal combined ontology. The assumptions of the reasoning model could then be made explicit. It also allows reasoning about the assumptions made by the data producer and the modeller, and may make it possible, for example, to retract the least plausible assumption(s) in the case of inconsistent data, using standard theories of belief revision, non-monotonic reasoning, or other uncertainty formalisms.

Formally if the product ontology can be expressed in terms of one logical language with meaning defined in terms of set of possible worlds $\mathcal{M}$, and the problem ontology in another language with set of possible worlds $\mathcal{M}'$, then the relationship between the two languages is determined by a relation between $\mathcal{M}$ and $\mathcal{M}'$. This allows imprecision in the translation between the ontologies. Furthermore, different assumptions about correspondences will lead to different such relations. These assumptions may be labelled simply with terms such as 'very reliable', 'reliable'; 'tentative', with totally ordered grades such as these, we get a 'fuzzy' relation between $\mathcal{M}$ and $\mathcal{M}'$, and it can be viewed in terms of possibility theory. Alternatively, a more sophisticated lattice-based representation for grades might be used, allowing grades of quality to be partially-ordered (e.g., if we do not know which assumption is more reliable).

The present status of the project is illustrated by experiments on real scale applications.

## 2. The place of ontologies in geographic information

In the context of the European project REV!GIS, we are faced with recurrent problems raised by many geographic applications, such as:

-the multi-sources integration problem;

-the land cover change detection problem;

-the extrapolation of incomplete data under propagation constraints.

Actually these are subcases of the "data fusion" problem, as it is often named, and actually, it is a particular case of the "knowledge fusion" problem, as it should be named if it is studied from an artificial intelligence viewpoint (see: [1]).

Let's detail further this fusion problem: the common background is that we are looking at a same portion of the geographical space (a region), with the help of two (or more, but two are sufficient for our purpose) datasets. Each dataset carries along its own "understanding" of this region, and the observations on which it is based are acquired at a certain scale level (both in space and time), and for a certain purpose in mind. These elements of understanding can be merged into a framework called an ontology (see: [2]).

Consequently any "perceived differences" between the datasets can have several causes:
(1) a perceived difference relates two "objects" of the region, that we can reasonably assume being the same entity of the Real World, and the perceived difference can an error on at least one dataset if we don't expect such difference: if one dataset is more trustworthy than the other, we can decide to "revise" this probable error ;

(2) a perceived difference relates two "objects" of the region, that we can reasonably assume being the same entity of the Real World, and if we assume also that there is a significant time gap between the two datasets, then the perceived difference denotes a true "change" during this gap;

(3) a perceived difference relates two "objects" which have not exactly the same definition, and we are perceiving this difference (see: [3]);

Let's name (1) the "*error*", (2) the "*change*", (3) the "*ontological difference*".

Very often, it is probably a mix of the three, but we expect always to have one cause being the most important.

Then, the above problems can be reformulated this way:
   - the multi-sources integration: we want two build a single dataset by keeping the "best" information from the two sources, hence we are seeking "true errors", which are neither change nor ontological differences ;
   - the land cover change detection: we want to collect all the "true changes", which are neither errors, nor ontological differences;
   - extrapolation under constraints: we want to fill the incomplete dataset by computing some new data which comply with the previous semantics of the original data and are consistant with the constraints.

Conclusion, the ability to discern ontological differences is mandatory for this kind of geographical applications.

The approach chosen by the REV!GIS consortium to handle this issue is to handle the underlying data management problem as an integration of both the data and their semantics, within a common reasoning framework. This is often refered to as an "interoperability" problem, but the literature in this field sometimes splits this problem into: "data interoperability" and "process interoperability". Our purpose is concerned by this distinction, and put the focus on a special topic, the "quality", which is rarely addressed in such designs.

Our "ontological approach" privilegies a separation between the data as a "product output", and the data as a "problem input", hence the use of two different kinds of ontologies: the " Product Ontologies " and the " Problem Ontologies ".

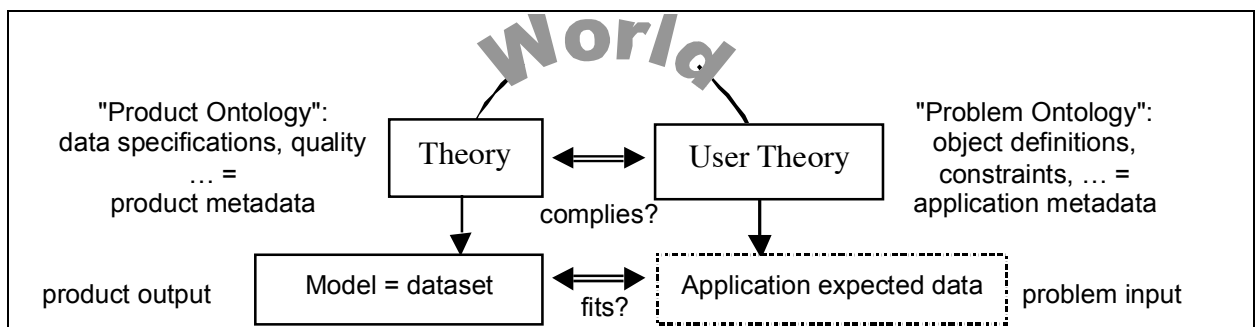

*Figure 1: ontological separation: two non-crossing visions of the World for the same data*

What we mean by " Product Ontologies " is any kind of information, and as many as possible, on all the concepts that a specific product carries:
   - explicitely available: sometimes the source metadata can provide a detailed and important part of it, sometimes it is present through various written documents ;

- implicitly: either it belongs to a more general ontology, not given because of its alleged obviousness, or simply it's forgotten.

We can predict that the collection of this information will be always incomplete, but we are presently trying to collect and formalize within a logical framework (a Prolog set of clauses) the "ontologies" of some very well documented datasets, provided by some major mapping agencies. Of course, the "quality information" is part of the product ontology. Examples are given in section 6.

What we mean by " <u>Problem Ontologies</u> " is all the "relevant information" for the problem raised in the context of the application. Of course the user cannot provide a priori all the knowledge he expects that will be relevant, but we assume that he should be able to say explicitly if some specified information is or is not relevant, or to some extent, and this "relevance grade" is part of the problem ontology. This describes the "virtual dataset" that the user would like to use, instead of the "actual dataset". Examples are given in section 6.

Finally, -and this is the core of the project REV!GIS-, the conflation of the two ontologies should help us to measure, or at least to roughly figure out, if the quality of the product fits with the problem relevance.

Conclusion: to answer the question of the "fitness for use" of datasets in applications, we need first to integrate ontologies, because the "fitness" is a semantical notion.

## 3.  The Spatial Extrapolation Problem

This problem has been studied through a simple but didactic application: the flooding of the river Herault, in 1994, such as reported and investigated by the Cemagref (see: [4]).

<u>data</u>: one digital map of the "parcels" (*First*), interpreted from aerial pictures and completed by terrain observations and measurements, gives the topography, the land structure delimited by physical limits which are relevant for the water flooding: hedges, drains … plus the "visibility of the vegetation" (even/over/under water), and the collection of the visible "flows" (*Second*), associated to the above limits ;

<u>processing</u>: the geometric registration and the terrain elevation are assumed to be perfect, but the translation of the visibility of the vegetation into interval bounds for the height of the water, is rather uncertain; on the contrary, the flow direction is certain, when visible, but this information is rather incomplete;

<u>problem</u>: let's combine the two products, and extrapolate into a "more" complete map: *Third*. How to perform this "extrapolation" ?

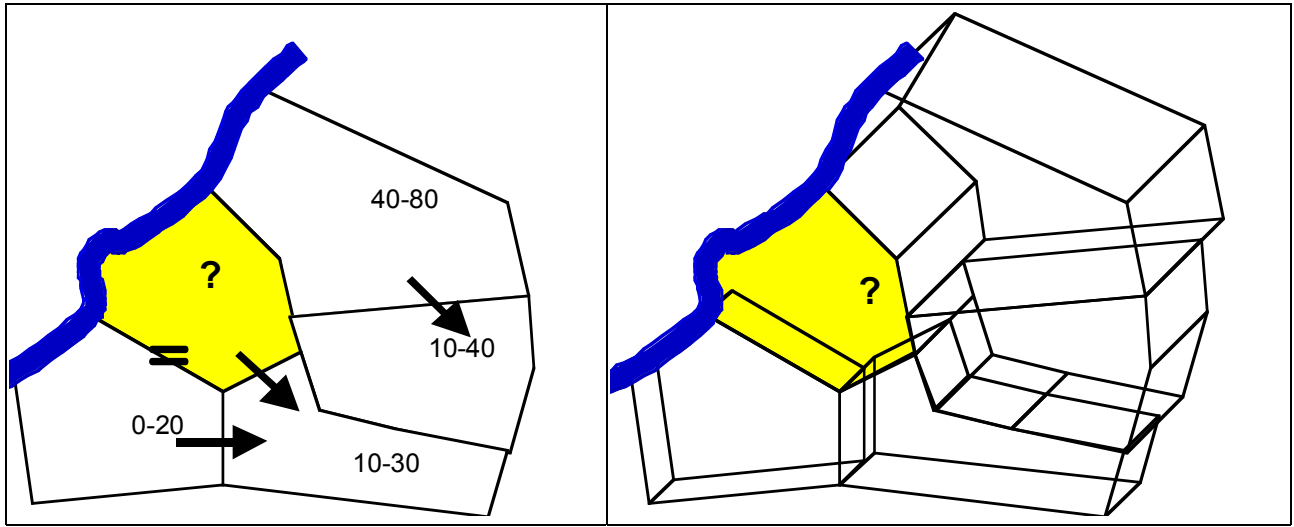

*Figure 2: the relationship between flows and water-height intervals, in the flooding application*

<u>The solutions</u>

(a) Numerical Interpolation:

   let's assume that we can observe (Figure 2: left) the vegetation-to-water relation for several parcels, and provide some min-max intervals assessments (right). We may try to interpolate, by some numerical computation, a possible value for the parcel with the question mark.
This interpolation takes into account the neighbouring intervals, but not the orientations of the known flows: a possible computed value may be: 15-40. We can say that a part of the application data (the flows) and of the associated knowledge (water flows down), is not used by the interpolation process.
AVOID THIS SOLUTION

(b) Interval Constraints Propagation:

   to combine (to merge) the two informations (known intervals and known flows) is a good means to compute missing values, while preserving the "water flows down" rule. The counterpart is that some original interval values may not comply the rule. Hence, what to do ? An experimentation based on this solution is presented in: [5].

   PREFER THIS SOLUTION, BUT: NEEDS CONSISTENCY CHECKING AND REVISION

## 4. The Spatial Data Fusion Problem

Let's consider a simple case of "data fusion":

<u>data</u>: two digital maps, from two different products or even two producers, of the same geographical region, one recording the streams (*First*), one recording the roads (*Second*) ;

<u>processing</u>: the geometric registration is assumed to be perfect, but the two kinds of features have different qualities, i.e. they depart differently from the "real features" ;

<u>problem</u>: let's combine the two products into a "more" informative map: *Third*. How to perform this "fusion" ?

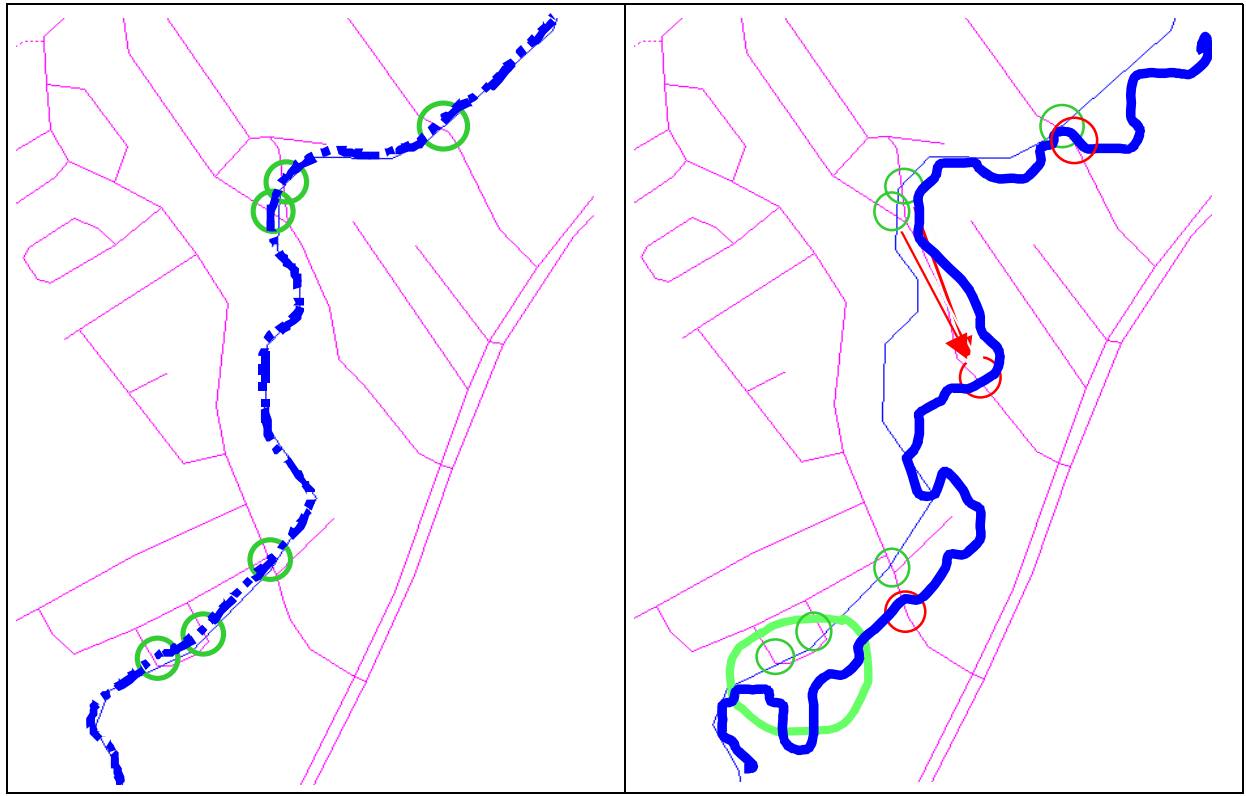

*Figure 3: the impact of the quality on the data fusion*

The solutions

(a) Set Theoretic Union:

To perform the mere 'set theoretic union' of the two datasets: each original one is strictly included in the union: is this order relation 'mappable' to the respective qualities ? The new map is expected to be better than the other two, but: Is its quality better everywhere ? How much it departs from the real features ?

Let's consider this simple constraint: « a road and a stream intersect at a bridge ». Hence there must be a bijection between bridges and intersections. Let v(i,j) be the property of the point (i,j), if: v(i,j) ∈ *First*, and v(i,j) ∈ *Second*, then this is an intersection point and the predicate "bridge(i,j)" must hold.

Now let's have a look to actual data: the left part of the Figure 3 is the mere union of streams (dot-dashed lines) and roads. The right figure overlays a new '*Second*' data (thick line), which is a different version of the same stream, supposedly more accurate (larger scale) and several problems can be noticed: we can list them and consider which quality parameters are addressed for each one, these "quality parameters" being chosen among the classical list of 5 (or 7) parameters present in the specialized literature.

--Displacement only modifications (top right): the associated quality parameter is "geometric accuracy";
--Complex modifications: two bridges are mixed into one, and several quality parameters may be associated, such as: commission error value for the "thematic accuracy", topological consistency value for the "logical consistency", and again the "geometric accuracy";
--Cardinality modifications (bottom), …

Depending on the user needs (contained in its ontology) some errors are more significant than other and this is what decides about the "fitness for use" of the data:
--from a navigation viewpoint, the topological consistency is the most important: eg. "*pass the bridge, then turn right*" relies fully on topology, and does not pay attention to the geometric accuracy;
--from a damage assessment viewpoint, the geometry can be more important: eg. "*all bridges within 1km range from the earthquake center must be controlled*".

Conclusion: the simple union of two dataset is a sound algebraic operation, but it does not reflect the implicit knowledge (semantical), associated to the data.
AVOID THIS SOLUTION

(b) Knowledge Fusion:

This second solution means that the information compounds both data and possibly inference rules, hence the logical closure of these rules. Therefore, the chosen set of rules will decide about the quality: eg. a simple coordinate shift can be ignored, but a difference in the total number of bridges can be a direct inconsistency.

The importance given to the rules by one user has a direct impact on the importance of one parameter of the quality (geometric accuracy, preservation of the topology, preservation of cardinalities …).

Conclusion: the "quality" is a semantical notion, not just a number or a probability, and it depends on the user needs and objectives. It is an "ontological issue". Moreover, the tasks of checking the inconsistencies, then to build possible solutions, and finally to find the "optimal" or "best" solution, are computationally difficult tasks (non polynomial).
PREFER THIS SOLUTION, EVEN IF IT IS HARDLY TRACTABLE (!)

## 5. The Land Cover Change Problem

Let's consider a simple case of "change detection":

data: two satellite images, from the same sensor, of the same geographical region but taken at two different dates ;

processing: the geometric registration is assumed to be perfect, but the images have been interpreted, pixel by pixel, according to two slightly different sets of classes: *First & Second* ;

problem: a same pixel (i,j) gives two values: v(i,j)=first $\in$ *First*, and v(i,j)= second $\in$ *Second*. How may we compare these two values ?

The solutions

(a) Numerical Difference:

all the values are numerical, for instance in the "radiometry" (reflectance or emitted energy) domain, and those numbers are "comparable" as such. There is no "semantics" at all, excepting the semantics of arithmetics, but the arithmetic operators have algebraic properties and we can compute anything … or nothing at all. Even the (numerical) equality has no semantics: eg. if first = second, does it means that there is "no change" ? Of course not, it may depend on several more assumptions. The same question raises with any distance such as the quantity | second-first |: what does it measure ? Easy to compute (just numbers), hard to answer (no semantics).

AVOID this solution.

(b) Contextual Comparison:

the sets *First* and *Second* are similar enough to be considered equal and renamed *Set*, hence the comparison between first(i,j) and second(i,j) depends on the existence of a comparison operator within this common *Set*. For instance, let's consider a possible "equal" operator, as a mere "string" comparator: if string( 'first' ) = string( 'second' ), then we can conclude that there is "no change" … excepting if: the assumption of the equivalence of *First* and *Second* is not fully acceptable, or if 'first' is dynamical by nature, such as a "cultivated land": does it means that there is no change in land cover or in land use ? (an agronom and a land surveyor will answer differently). About distances: the coding into 'first' and 'second' cannot lead directly to a distance computation, unless you provide it explicitely within a 2-dimension table, such as a "look-up table" (LUT). But what is the domain in which the distance holds: the one of the agronom, or the one of the land surveyor ? Again a semantics question.

BE CAREFUL with this solution.

(c) Ontological Comparison:

the sets *First* and *Second* are defined within two different ontologies (see: [6]) whose semantics are still similar but slightly different: the formal description of the two ontologies can help to understand these differences, and it already improves the previous solution (actually it is the way the "experts" follow, even without using the word ontology). But what is the status of the comparison: can we decide to perform it within *First* ?, or *Second* ?, or within a *Third* domain ? This last one is what we propose, because it subsumes the other two anyway and provides a formal description of a new ontology, to which *Third* belongs. In this solution, you know in which space you can build a distance.

PREFER this solution:
- it is more complex,
- it needs to explicit the differences between the original datasets and their respective interpretation processes,
- it needs to make the final intension explicit, by specifying the ontology in which the change detection must be assessed,

but
- it provides a full semantics to the change detection operator: *First* x *Second* → *Third* , either as a boolean or as a distance operator.

### The expert approach in the British "Land Cover Map" experiment:

The countryside survey (CS) has been performed in Great Britain in 1990, and was used for instance for deriving the Corine Land-Cover data. A second CS has been done in 2000, following several recommendations: some of these are easier to address than others, such as the timing of data capture. However, the other differences are fundamental to each technique and their integration may be irreconcilable, from the thematic perspective of land cover reporting. Examination of the definitions of the Broad Habitat classes used by each approach shows that the classes are described similarly, principally on their botanical and bio-geographic properties. Descriptions can be found at: http://www.ceh.ac.uk/data/lcm and http://www.cs2000.org.uk/ .

This problem of relating data of different pedigrees has been described and analysed in much work. The various solutions have in common some form of translation mechanism between the concepts from one dataset to those of another. The origins of the translation obviously depend upon the context (national, agency, disciplinary) in which the designer constructs the ontologies.

We are concerned with the general problem of revision, because we have to construct and then revise a translation between different conceptualisations (belief revision). It is expected that for some object, by their (physical) nature, the translation from Dataset1 to its as yet unknown equivalent classes in Dataset2, will be less problematic for some classes. In this sense the translation of such a class requires less revision than other classes. We illustrate an outline approach for this revision, using an example from the land cover classifications (see: Table 1 to Table 4 in Annexes).

Dataset1: LCMGB90 `http://www.ceh.ac.uk/data/lcm/lcmleaflet2000/leaflet3.pdf`
– Coniferous / evergreen woodland (label '16' in the 25 'target' class 1990 dataset).
– Bog (herbaceous): Lowland bog ( label '24')
– Bog (herbaceous): Upland bog (label '17').

Dataset2: LCM2000
- Coniferous woodland (alpha-code: C)
- Bog (alpha-code: B)

In terms of Metadata, we can record a 'strong' value for the level 'class quality' when its class value is 'coniferous' and a 'weak' for 'bog'.

## 6. Conclusion

The most important key word is: "semantics" ! We do need to take the semantics into account, even if many operations can be performed on a computer, at a simple algebraic level. This is obvious in particular with raster data, where the pixels can be used as numerical inputs of an infinity of simple to complex models ! It is not enough to try to make sophisticated models, the semantics of the data must be present at a higher level: before calling such or such numerical model (and then you know why this one, what are its limits, …), and when considering the results (how they relate really to the expected variable, how the quality of the computation, and of any preprocessing, impacts on the fitness of this result for its use, …). The exclamation mark in the name of the project REV!GIS has the same meaning as the conclusion of this paper: beware of the semantics of the data and you will be able to interpret their quality for your purpose !

## 7. References

[1] Papini, O. (2000). « Knowledge-base revision ». *The Knowledge Engineering Review*. Cambridge University Press, Volume **15**, numero 4. pages 339-370.

[2] Bittner, T.E. & G. Edwards, (2001). « Towards an Ontology for Geomatics », Geomatica, Journal of the Canadian Institute of Geomatics, vol. **55**, No. 4, 2001, p. 475-490

[3] Brodeur, J., Y. Bédard, B. Moulin & G. Edwards, (2001). « Geosemantics proximity and data fusion », GIFaR'01: International Workshop Geographic Information Fusion and Revision, April 2001, Québec, Canada.

[4] Puech, C., Raclot, D., 2002. Using GIS and aerial photographs to determine the water levels during flood. "Remote sensing and hydrology". Accepted in "Hydrological Processes".

[5] Würbel E., Jeansoulin, R., & Papini, O. (2000). « Revision : An application in the framework of GIS », Proceedings of the Seventh Internationnal Conference about Principles of Knowledge Representation and Reasoning, KR'2000,( April 2000), Anthony G. Cohn and Fausto Giunchiglia and Bart Selman ed, KR, inc., Morgan Kaufmann. Breckenridge, Colorado, USA, pp. 505-516.

[6] Comber, A.J. & Fisher, P.F. & Wadsworth, R. (2002). « Creating Spatial Information: Commissioning the UK Land Cover Map 2000 », Spatial Data Handling, SDH'02, Ottawa, August 2002.

**Acknowledgements**

This work is supported by the IST-FET ("Future & Emerging Technologies") programme of the European Community (5[th] Framework Program). The following universities, institutes or companies, collaborate to this work: Université de Provence, University of Leicester, Keele University, Université Laval, Université de Toulon, Technical University of Vienna, ITC, CNUCE, Somei-Praxitec.

ANNEX1: Land Cover Mapping classes for the years 1990, then 2000 (see: [Fisher])

| LCMGB90 | "key classes" (17): 1x1 km | # | "target classes" (25): 25x25 m |
|---|---|---|---|
| Sea | Sea / Estuary | 1 | Sea / Estuary |
| Inland water | Inland Water | 2 | Inland Water |
| Littoral | Beach / Mudflat / Cliffs | 3 | Beach and Coastal Bare |
|  | Saltmarsh | 4 | Saltmarsh |
| ? | Rough pasture, grass moor, … | 5 | Grass Heath |
|  |  | 9 | Moorland Grass |
|  | Pasture, meadow, amenity grass | 6 | Mown / Grazed Turf |
|  |  | 7 | Meadow / Verge / Semi-natural |
|  | Marsh / Rough Grass | 19 | Ruderal Weed |
|  |  | 23 | Felled Forest |
|  |  | 8 | Rough / Marsh Grass |
|  | Grass Shrub Heath | 25 | Open Shrub Heath |
|  |  | 10 | Open Shrub Moor |
|  | Shrub Heath | 13 | Dense Shrub Heath |
|  |  | 11 | Dense Shrub Moor |
|  | Bracken | 12 | Bracken |
| Mixed woodland | Deciduous / Mixed Wood | 14 | Scrub / Orchard |
|  |  | 15 | Deciduous Woodland |
| Coniferous | Coniferous / Evergreen Woodland | 16 | Coniferous Woodland |
| Bog | Bog (Herbaceous) | 24 | Lowland Bog |
|  |  | 17 | Upland Bog |
| Arable | Tilled (Arable Crops) | 18 | Tilled Land |
| Built-up | Suburban / Rural Development | 20 | Suburban / Rural Development |
|  | Urban Development | 21 | Continuous Urban |
| Bare | Inland Bare Ground | 22 | Inland Bare Ground |

*Table 1: LCMGB 1990 classes*

| LCM2000 Top Level | LCM Target class | LCM Subclass (Lev2) | Alpha |
|---|---|---|---|
| Sea | Sea / Estuary | Sea / Estuary | We |
| Water inland | Water (inland) | Water (inland) | W |
| Littoral sediments | Littoral rock sediments | Littoral rock | Lr |
|  |  | Littoral sediment | Lm |
|  |  | Saltmarsh | Lsm |
| Supra littoral sediments | Supra-littoral rock sediments | Supra-littoral rock | Sr |
|  | sediment | Supra-littoral sediment | Sh |
| Bog | Bog | Bogs (deep peat) | Bh |
|  | Dwarf shrub heath | Dense dwarf shrub h. | H |
|  |  | Open dwarf shrub h. | Hga |
|  | Montane habitats | Montane habitats | Z |
| Mixed woodland | Broad mix woodland | Broad mixed woodland | D |
| Coniferous | Coniferous woodland | Coniferous woodland | C |
| Arable | Arable and hortic | Arable cereals | Ab |
|  |  | Arable horticulture | Aba |
|  |  | Non-rotational arable | Ado |
|  |  | and horticulture | Agl |
| Grassland | Improved grassland | Improved grassland | Gi |
|  | Rough and semi-nat | Setaside grass | Gis |
|  | neutral and calcare | Neutral grass | Grn |
|  | grasslands | Calcareous grass | Gc |
|  | Acid grass and brac | Acid grass | Ga |
|  |  | Bracken | Gbr |
|  | Fen, marsh sw | Fen, marsh & swamp | Fs |
| Built-up | Built up areas gard. | Suburban/rural devlpd | Us |
|  |  | Continuous Urban | U |
| Bare | Inland Bare Ground | Inland Bare Ground | Id |

*Table 2: LCM2000 Top classes (11), Target (16), Subclasses (26) and Variants (72), with alpha codes*

ANNEX2: summary of classifications

(Note1: the word "target" corresponds to two different levels in the two tables: target90 = ± subclass2k )
(Note2: the "top" class in both tables is not provided by the CEH experts, but as a tentative to form a group at a coarser level)

Excerpt from the entire LCMGB90 classification:

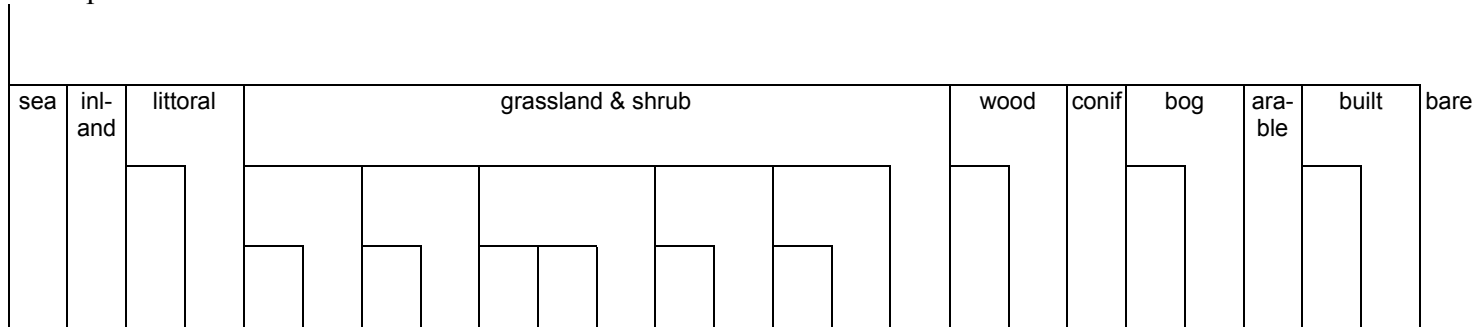

*Table 1: hierarchy of top classes (10), key classes (17), and targets (25)*

Excerpt from the entire LCM2000 classification:

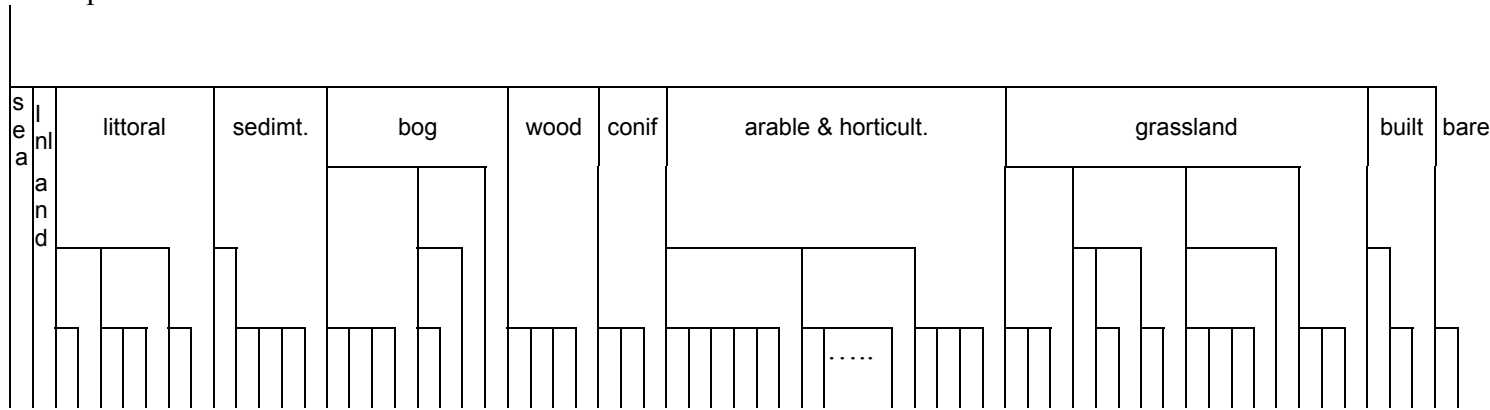

*Table 2: hierarchy of top classes (11), targets (16), subclasses (26) and variants (72)*